\documentclass[letterpaper, 10 pt, conference]{ieeeconf}  

\IEEEoverridecommandlockouts                              

\usepackage[table,xcdraw]{xcolor}
\usepackage{bbding}
\usepackage{graphicx}
\usepackage{stfloats}
\usepackage{float}
\usepackage{booktabs}
\usepackage{textalpha}
\usepackage{subcaption}
\usepackage{amssymb}
\usepackage{amsmath}
\usepackage{multirow}
\usepackage{verbatim}

\title{\LARGE \bf
EVEN: An Event-Based Framework for Monocular Depth Estimation at Adverse Night Conditions
}

\author{Peilun Shi$^{1}$, Jiachuan Peng$^{1}$, Jianing Qiu$^{1,2,\dagger}$, Xinwei Ju$^{1}$, Frank Po Wen Lo$^{1}$, and Benny Lo$^{1}$
\thanks{$^{\dagger}$indicates the corresponding author}
\thanks{$^{1}$The Hamlyn Centre, Imperial College London, London, U.K. 
        {\tt\small \{p.shi21, j.peng21, x.ju21, po.lo15, benny.lo\}@imperial.ac.uk}}%
\thanks{$^{2}$Department of Computing, Imperial College London, London, U.K. 
        {\tt\small jianing.qiu17@imperial.ac.uk}}%
\thanks{The authors would like to thank Dr Lin Wang from Hong Kong University of Science and Technology, Guangzhou for his helpful discussion.}
}

\begin{document}

\maketitle
\thispagestyle{empty}
\pagestyle{empty}

\begin{abstract}

Accurate depth estimation under adverse night conditions has practical impact and applications, such as on autonomous driving and rescue robots. In this work, we studied monocular depth estimation at night time in which various adverse weather, light, and different road conditions exist, with data captured in both RGB and event modalities. Event camera can better capture intensity changes by virtue of its high dynamic range (HDR), which is particularly suitable to be applied at adverse night conditions in which the amount of light is limited in the scene. Although event data can retain visual perception that conventional RGB camera may fail to capture, the lack of texture and color information of event data hinders its applicability to accurately estimate depth alone. To tackle this problem, we propose an event-vision based framework that integrates low-light enhancement for the RGB source, and exploits the complementary merits of RGB and event data. A dataset that includes paired RGB and event streams, and ground truth depth maps has been constructed. Comprehensive experiments have been conducted, and the impact of different adverse weather combinations on the performance of framework has also been investigated. The results have shown that our proposed framework can better estimate monocular depth at adverse nights than six baselines.

\end{abstract}

\section{Introduction}\label{sec:intro}

\begin{figure}[h]
\centerline{\includegraphics[width=\linewidth,scale=1.0]{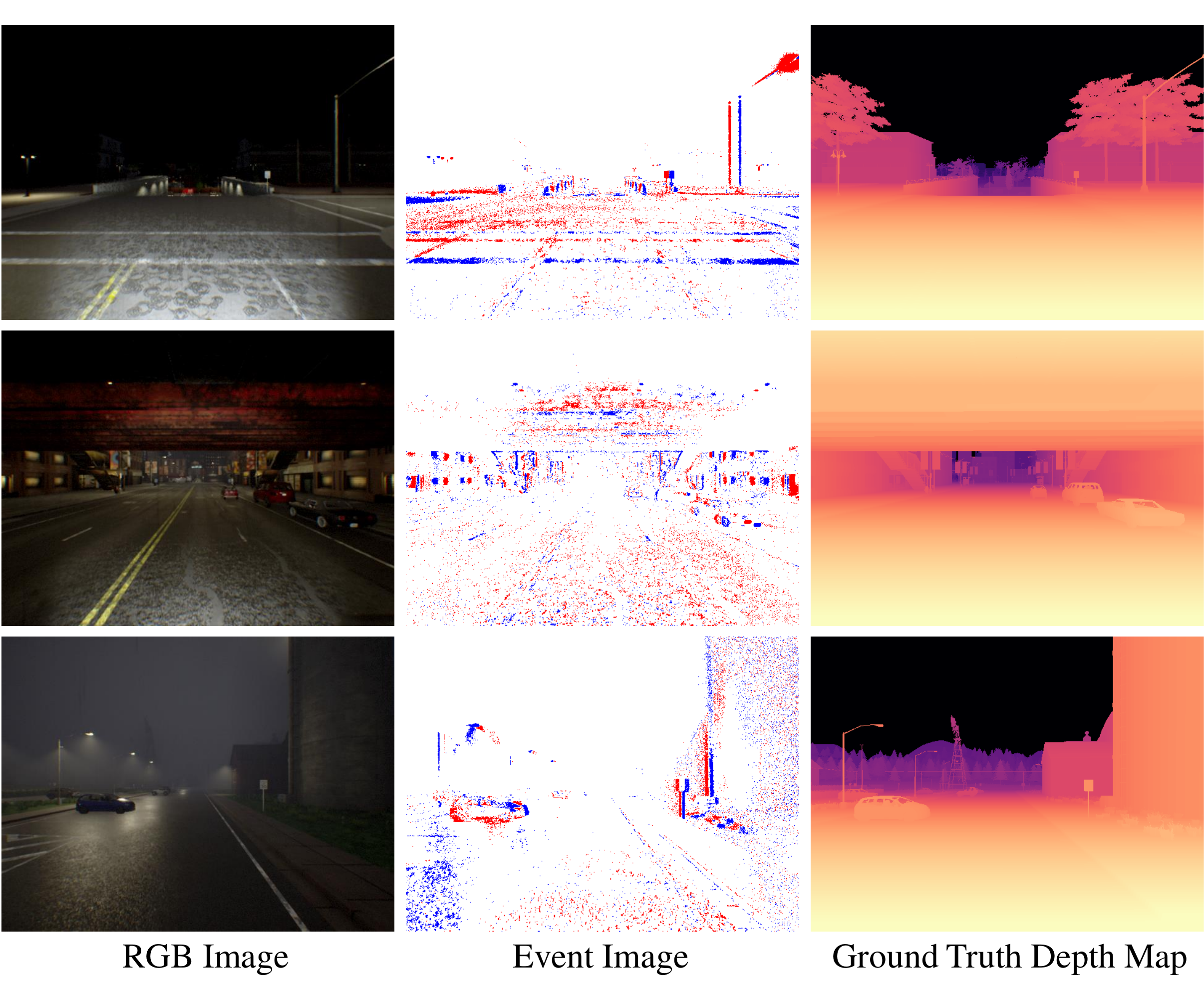}}
\caption{Data samples from our MonoANC dataset. We show paired RGB and event images, and the ground truth depth map for each sample. The adverse night scenarios from top to bottom are: 1) driving in the heavy rain on a city road; 2) driving under a bridge at a foggy night; 3) driving at the countryside at a rainy and foggy night.}
\label{fig:framework}
\end{figure}

Depth estimation with monocular cameras has been actively studied over the past decades~\cite{Godard_2017_CVPR, godard2019digging, casser2019depth}, as it offers an efficient and economic way of obtaining depth. Compared to LiDAR, a monocular camera can be deployed pervasively, and due to its small scale, it can also be installed on an agent, e.g., an autonomous car, unobtrusively. 

Albeit convenient and flexible, accurately estimating depth from a monocular camera is non-trivial, especially at night time, at which the visual perception of conventional RGB cameras degrades. The low dynamic range and sensitivity to motion blur of conventional cameras can lead to defective imaging at night, and the captured images/videos often exhibit underexposure due to low-lighting or back-lighting~\cite{wang2019underexposed}. For an autonomous car, when it is driving at night accompanied by adverse weather (e.g., rain and fog), the dual occurrence of adverse light and weather can cause a challenge for its RGB-based vision system.

Recently, event camera has gained popularity in visual perception and robotics. Event camera is a bio-inspired vision sensor that works in a different way than conventional cameras \cite{gallego2020event, lichtsteiner2008c}. Rather than capturing intensity images at a fixed rate, event cameras measure intensity changes asynchronously in the form of an event stream. Event cameras have distinct advantages over conventional RGB cameras, including very high dynamic range (HDR), high temporal resolution, less motion blur, and low power consumption. These features of the event camera can complement its RGB counterpart, providing extra visibility and leading to an enhanced visual perception system.

On the other hand, in depth estimation, texture and salient edges play more important roles than color as recognized by research in the computer vision community~\cite{hu2019visualization}. Texture can be well retained in RGB data whereas salient edges can be better captured by the event camera. Therefore, using both data modalities is a straightforward attempt to boost the overall depth estimation accuracy.

Although there are few studies ~\cite{gehrig2021combining},~\cite{Zhu_2018_ECCV},~\cite{tulyakov2019learning} that have been proposed to jointly utilize RGB and event data for monocular depth estimation, they mainly focus on day time or normal weather conditions. Thus far, no research has been carried out on event-based monocular depth estimation under adverse night conditions, which is challenging as the RGB source does not contain as much effective visual information as it does at day time, and how to effectively fuse RGB data with event stream at night time has yet to be addressed.

Despite practical applications, such as more intelligent and lightweight night-time autonomous driving and rescue robots, there is currently also no dataset that contains paired RGB, event and ground truth depth data captured at adverse night conditions to validate and benchmark research in this direction. Hence, in this work, we made the following two contributions:

\begin{enumerate}
    \item We propose the first adverse night-time driving dataset that contains paired RGB images, event streams, and ground truth depth maps. The adverse night conditions in our dataset are diverse in a variety of aspects including adverse weather such as rain and fog, and different scenes such as driving on dim countryside roads. 
    \item We propose a novel three-phase framework, which employs low-light enhancement and multi-modal fusion to tackle the problem of monocular depth estimation at adverse night conditions with event-based vision. The entire framework has been thoroughly evaluated, with the results showing that it outperforms six baselines.
\end{enumerate}

\section{Related Work}\label{sec:related_work}

\subsection{Monocular Depth Estimation with Multi-Modal Fusion}

Monocular depth estimation can be achieved using RGB modality alone~\cite{Godard_2017_CVPR, godard2019digging, casser2019depth}. Recent advances in multiple data modalities have further improved the depth estimation accuracy. For instance, some research works proposed to use RGB and optical flow~\cite{ranftl2016dense,tateno2017cnn,chen2020denao,shimada2022pix2pix}, RGB combined with segmentation maps~\cite{zama2018geometry, zhu2020edge, he2021sosd}, or RGB with extra saliency features~\cite{zhang2021deep, abdulwahab2022monocular} as the inputs, and use multi-modal fusion to enhance depth estimation. 

LiDAR has been explored for enhancing monocular depth estimation recently. \cite{jaritz2018sparse} and \cite{fu2019lidar} proposed using late fusion methods to fuse depth data from LiDAR and monocular RGB inputs. Apart from pure visual signals, radar has also been used with RGB modality for monocular depth estimation \cite{lin2020depth, lo2021depth}. Recently, an attention-based method has been proposed for fusing radar signals with monocular RGB images \cite{long2022radar}.

\subsection{Event-Based Depth Estimation}
Daniel et al.~\cite{gehrig2021combining} combined event-based data and monocular RGB frames with a recurrent asynchronous network for depth estimation, which is also the first work to fuse the event and monocular RGB frames. Zhou et al.~\cite{zhou2018semi} investigated the use of stereo event cameras for semi-dense depth estimation by
maximizing a temporal consistency between the corresponding event streams. Another event vision-based method was proposed by Zhu et al.~\cite{Zhu_2018_ECCV} which eliminates disparity for depth estimation. The method proposed by~\cite{tulyakov2019learning} shows the first learning-based stereo depth estimation for event cameras which is also the first one that produces dense results. \cite{zhu2019unsupervised} is an unsupervised framework that learns motion information only from event streams, achieving multi-task objectives including optical flow, egomotion and depth estimation. Cui et al. \cite{cui2022dense} proposed a dense depth estimation method based on the fusion of dense event stream and sparse point cloud.

Despite the efforts being made in event-based depth estimation, existing works are not engineered to specifically tackle monocular depth estimation at adverse night conditions, but instead mainly target at day time and normal weather conditions. In this work, we target monocular depth estimation at adverse night conditions. In order to improve the illumination in the field of view (FOV) and to take advantage of the HDR property of the event-based camera, we propose to combine low-light enhancement and multi-modal fusion of event and RGB data for better depth estimation. To the best of our knowledge, we are the first work that uses the event-based vision along with low-light image enhancement to estimate monocular depth at adverse night conditions.

\section{Method}\label{sec:method}

\begin{figure*}
    \centerline{\includegraphics[width=\textwidth,scale=1.0]{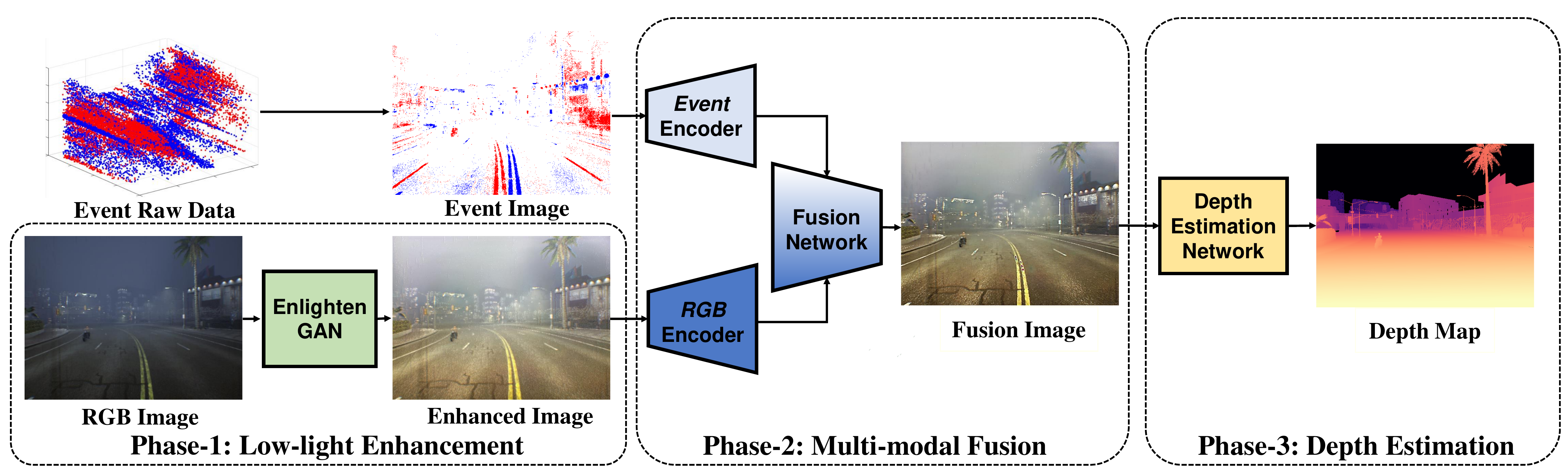}}
\caption{An overview of the proposed framework for monocular depth estimation at adverse night conditions (e.g., at foggy night). Our framework, named EVEN, leverages a three-phase process to estimate depth: 1) phase-1: enlightening the low-light RGB image; 2) phase-2: fusing visual information from enhanced RGB and event images; 3) phase-3: estimating depth based on reconstructed fusion image.}
\label{fig:framework_overview}
\end{figure*}

Our framework decomposes monocular depth estimation at adverse night conditions into three phases as shown in Fig.~\ref{fig:framework_overview}. In phase one, the raw RGB image is first enlightened using low-light image enhancement; In phase two, the enhanced RGB image and the event image are fused to generate a fusion image; In phase three, depth estimation is carried out based on the fusion image. We denote our framework as \textbf{EVEN} as it is based on \textbf{EV}ent vision and low-light \textbf{EN}hancement. We elaborate our framework in the following.

\begin{figure} 
\centerline{\includegraphics[width=\columnwidth,scale=1.0]{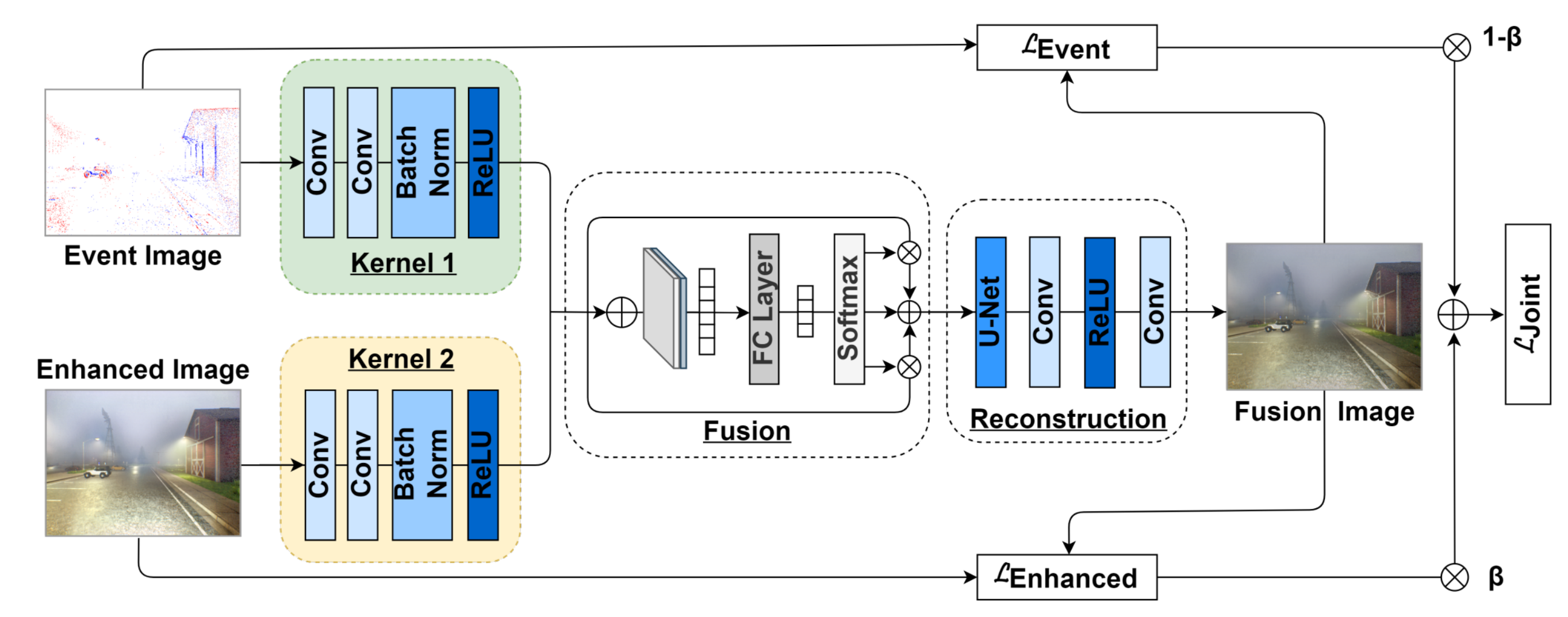}} 
\caption{The multi-modal fusion network of EVEN. }
\label{fig:fusion_network} 
\end{figure}

\subsection{Event Stream}\label{subsec: event_stream}
Asynchronous event streams reflect changes in light intensity. In order to efficiently make full use of the information from the event-based data. We convert event streams in the voxel grid format to image format. Specifically, spatial points (indexed by $x$ and $y$ positions in image coordinates with the value being the polarity $p$) are stacked along the time axis $t$ using a fixed time period $\Delta$$t$ = 0.125 s. This produces a compact event image.

\subsection{Phase-1: Low-light Enhancement}

The visual perception of conventional RGB cameras degrades at night due to the limited amount of light. To recover the necessary scene color and texture information captured by the RGB camera, we utilize EnlightenGAN \cite{jiang2021enlightengan} to enhance the raw night-time RGB image. EnlightenGAN is of an attention-based U-Net structure. The input RGB image is normalized by using the illumination channel as the attention map for the ultimate low-light enhancement.

\subsection{Phase-2: Multi-modal Fusion}

Event data can capture much more HDR and temporal details of night-time scenes, whereas RGB data can provide necessary texture and color information. As these two modalities complement each other, and in order to leverage the merits of both, a novel fusion network (refer to Fig.~\ref{fig:fusion_network}), which is built on top of selective kernel network~\cite{li2019selective}, is designed to integrate event data with RGB modality.

\subsubsection{Fusion Network}

given an event image $\mathbf{X}_{Event}$ and an enhanced RGB image $\mathbf{X}_{Enhanced}$, we use two convolutional kernels with different kernel sizes to transform the input images into feature maps. After transformation, two feature maps $\mathbf{F}_{Event}$ and $\mathbf{F}_{Enhanced}$ are obtained:

\begin{equation}
    \mathbf{F}_{Event} = g(\mathbf{X}_{Event}), \mathbf{F}_{Event} \in \mathbb{R}^{H \times W \times C}
\end{equation}

\begin{equation}
    \mathbf{F}_{Enhanced} = h(\mathbf{X}_{Enhanced}), \mathbf{F}_{Enhanced} \in \mathbb{R}^{H \times W \times C}
\end{equation}
where $g(\cdot)$ and $h(\cdot)$ are separate convolutional neural network layers that conduct transformation. For the event image, we use a kernel size of $5 \times 5$ as the information carried in event modality is relatively sparse. Therefore, a large kernel size is used. For the enhanced RGB image, we use a kernel size of $3 \times 3$. Following convolutional transformation, the feature maps of the two modalities are merged using an element-wise summation:

\begin{equation}
    \mathbf{F}_{sum} = \mathbf{F}_{Event} + \mathbf{F}_{Enhanced}, \mathbf{F}_{sum} \in \mathbb{R}^{H \times W \times C}
\end{equation}

We then apply global average pooling to conduct dimension reduction (along the $H$ and $W$ dimensions) for the merged feature map $\mathbf{F}_{sum}$, which produces a vector $\mathbf{V} \in \mathbb{R}^{1 \times C}$. Similar to~\cite{li2019selective}, we then use a simple fully connected layer $f(\cdot)$ to create a compact vector $\mathbf{k}$ on the basis of $\mathbf{V}$:

\begin{equation}
    \mathbf{k} = f(\mathbf{V}), \mathbf{k} \in \mathbb{R}^{d \times 1}
\end{equation}

$\mathbf{k}$ is then used to guide adaptive fusion of the two modalities. Specifically, we create soft attention across channel $C$. For $c$-th element along the channel $C$, the soft attention for fusing event and enhanced RGB feature maps can be formulated as follows:

\begin{equation}
a_c=\frac{e^{\mathbf{A}_c \mathbf{k}}}{e^{\mathbf{A}_c \mathbf{k}}+e^{\mathbf{B}_c \mathbf{k}}}, b_c=\frac{e^{\mathbf{B}_c \mathbf{k}}}{e^{\mathbf{A}_c \mathbf{k}}+e^{\mathbf{B}_c \mathbf{k}}}
\end{equation}

\begin{equation}
\mathbf{F}_{{fused}_c} = a_c \cdot \mathbf{F}_{{Event}_c}+b_c \cdot \mathbf{F}_{{Enhanced}_c}, a_c + b_c = 1
\end{equation}
where $\mathbf{A}_c \in \mathbb{R}^{1 \times d}$ and $\mathbf{B}_c  \in \mathbb{R}^{1 \times d}$ are learnable vectors.

The fused feature map $\mathbf{F}_{fused}$ is then fed into an U-Net~\cite{ronneberger2015u} followed by a group of convolution and ReLU operations to 1) further fuse features of the event and RGB modalities, and 2) reconstruct a fusion image $\mathbf{Y}$ of the same resolution to the input event and enhanced RGB images:

\begin{equation}
    \mathbf{Y} = \text{Conv}(\text{ReLU}(\text{Conv}(\text{U-Net}(\mathbf{F}_{fused}))))
\end{equation}

The resulting fusion image, which has HDR property and better edge salience, also suppresses areas of overexposure caused by low-light enhancement as shown in Fig.~\ref{fig:fusion_network}.

\subsubsection{Fusion Loss}\label{AA}
In order to allow the entire fusion network to effectively merge visual information from the two modalities, a joint loss $\mathcal{L}_{joint}$ is designed as shown in Equation~\ref{eq:joint_loss}. We use the reconstruction loss between the fusion image and the enhanced RGB image as the primary loss (i.e., $\mathcal{L}_{Enhanced}$), and that between the fusion image and the event image as the auxiliary loss (i.e., $\mathcal{L}_{Event}$). Both reconstruction losses are implemented as an $\mathcal{L}_2$ loss that measures the mean squared error between the fusion image and the respective event or enhanced RGB image. During training, the fusion network is trained to decrease $\mathcal{L}_{joint}$.

\begin{equation}\label{eq:joint_loss}
    \mathcal{L}_{joint} = \beta \times \mathcal{L}_{Enhanced} + (1 - \beta) \times \mathcal{L}_{Event}
\end{equation}

\subsection{Phase-3: Depth Estimation}\label{AA}
The fusion image, which contains visual information from both event and RGB modalities, is then used as the source for depth estimation. We separately adopt two state-of-the-art depth estimation networks, i.e., Depthformer~\cite{li2022depthformer} and SimIPU~\cite{li2021simipu} in our EVEN framework to carry out the depth estimation with the fusion image as their input.

\section{Dataset}\label{sec:dataset}
To the best of our knowledge, there is currently no dataset that is proposed for monocular depth estimation at adverse night conditions, containing paired RGB, event and depth images. In order to validate the effectiveness of our proposed framework, and advance future research in this direction, we construct the first adverse night-time driving dataset that includes the aforementioned data modalities and the ground truth depth maps. The dataset was constructed using CARLA \cite{dosovitskiy2017carla}, a popular simulator in the field of autonomous driving, and the event camera plugin \cite{hidalgo2020learning}.

\subsection{Data Collection and Statistics}
\begin{figure}
\centering
\begin{subfigure}{.5\columnwidth}
  \centering
  \includegraphics[width=\columnwidth]{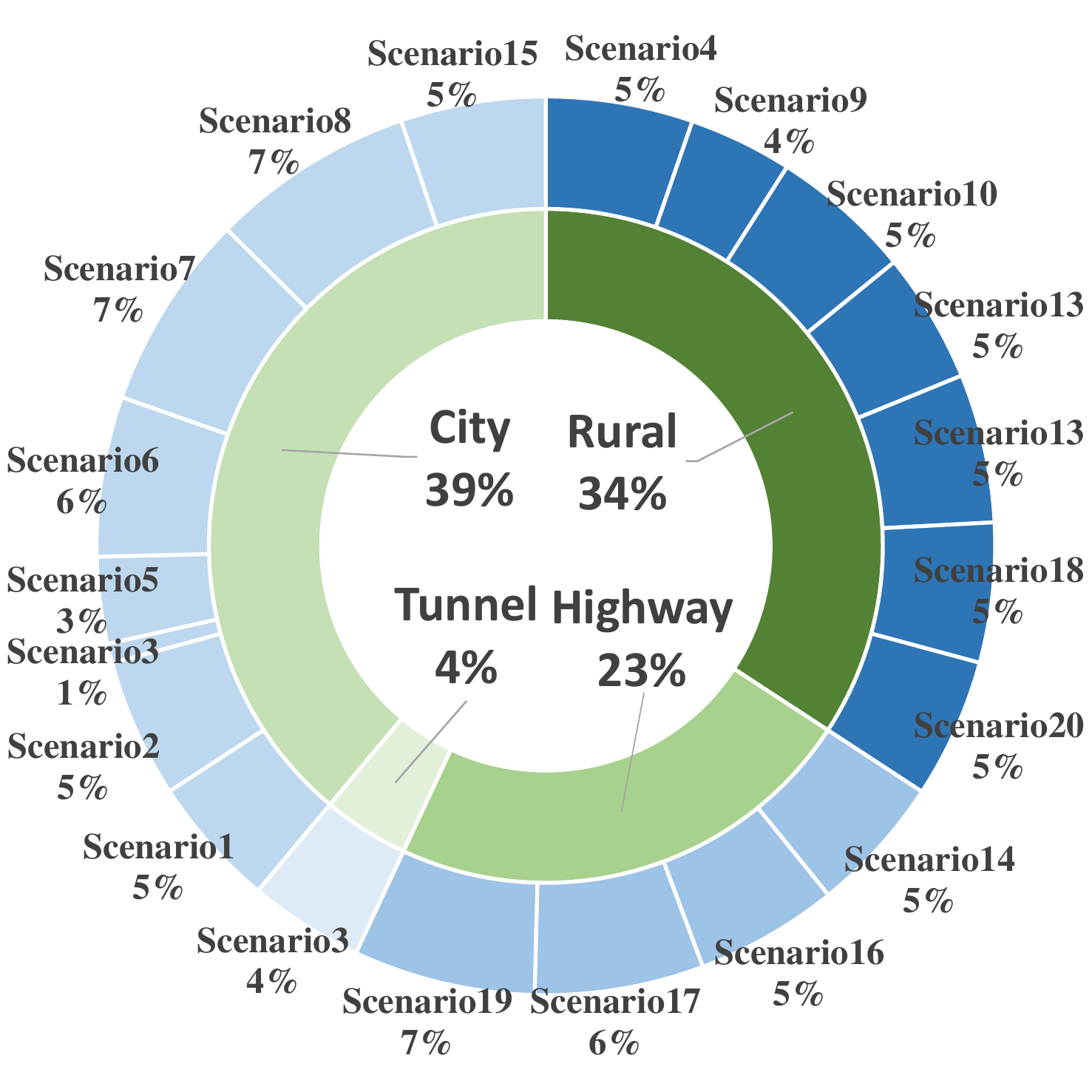}
  \caption{}
  \label{fig:scene_distribution}
\end{subfigure}%
\begin{subfigure}{.5\columnwidth}
  \centering
  \includegraphics[width=\columnwidth]{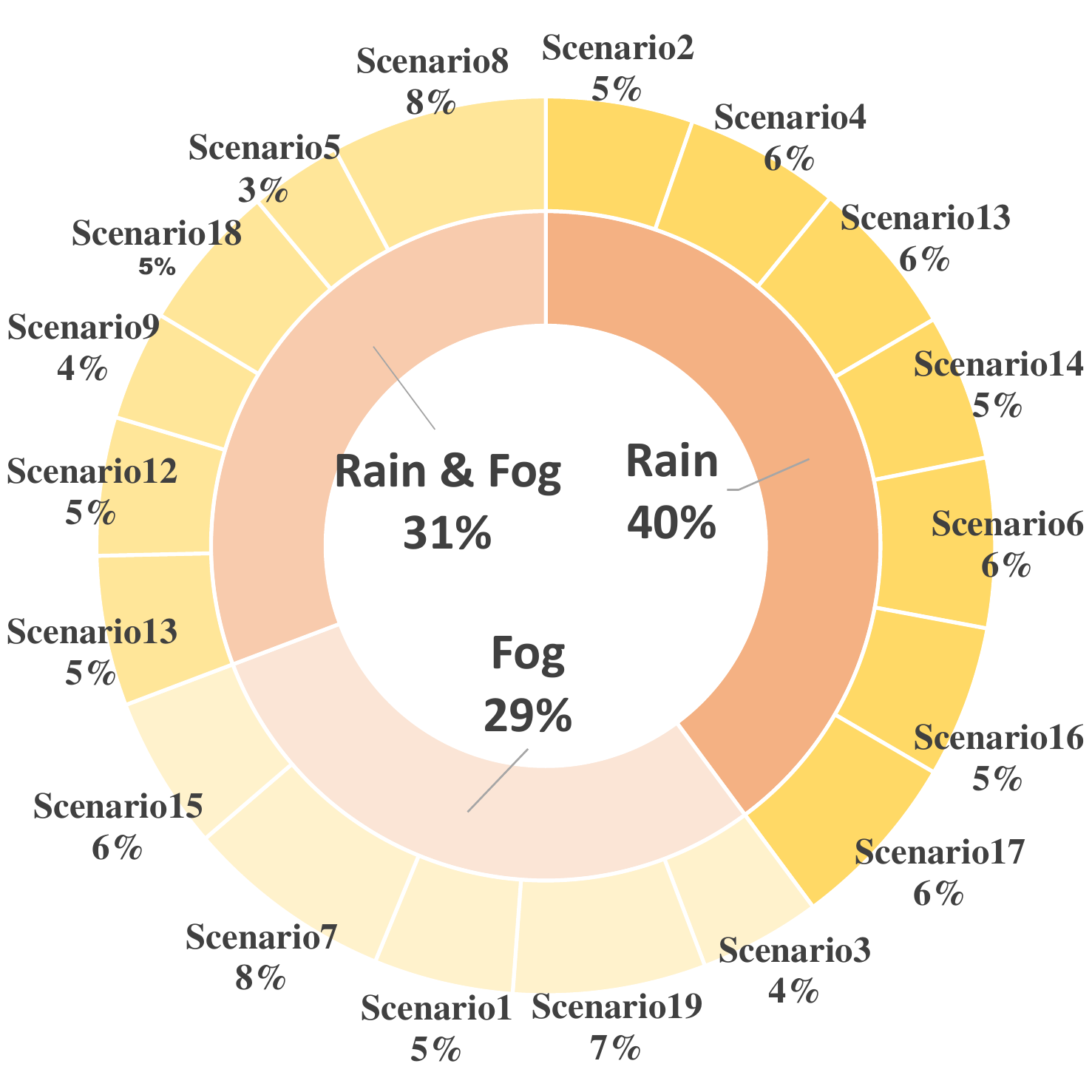}
  \caption{}
  \label{fig:weather_distribution}
\end{subfigure}%
\caption{Distribution of different night-time driving environments (a) and different adverse weather conditions (b).}
\label{fig:dataset_statistics}
\end{figure}

We collect the data through a sensor suite that contains an RGB camera, an event-based camera, and a depth camera. The positive and negative thresholds for triggering an event of the event camera was set to 0.4. All the sensors were set with a FOV of 90 degrees, a resolution of $640 \times 480$ pixels, and a data generation rate of 8 Hz. All sensors had an egocentric view mounted on a vehicle while it was driving around. The start and end points of the vehicle were manually selected and the routes of the vehicle were accurately planned. The speed and following distance of the vehicles as well as the lights at night were based on road traffic guidelines to achieve the maximum realism of night-time driving. The driving scenes are diverse, including typical city and rural roads, as well as tunnel and highway. The statistics of the driving scenarios is shown in Fig.~\ref{fig:scene_distribution}. We also apply adverse weather conditions such as rain, fog, and the combination of rain and fog to the scene, and Fig.~\ref{fig:weather_distribution} shows the distribution of the adverse weather in our dataset. The dataset contains 11,191 samples. Each sample has paired RGB, event and ground truth depth images. We split the entire dataset into 70\% for training, 15\% for validation and the rest 15\% for testing. We name our dataset as \textbf{MonoANC} (\textbf{Mono}cular depth estimation at \textbf{A}dverse \textbf{N}ight \textbf{C}onditions).

\section{Experiment}\label{sec:experiment}
In this section, we first describe the implementation details of our framework - EVEN, and then the evaluation metrics, followed by the baseline methods that are used to compare against our framework. We then show overall results of all methods on MonoANC, and present the results of cross validation of the performance of EVEN on different adverse weather combinations at the end.

\subsection{Implementation Details}

We implement our EVEN framework using PyTorch. The learning rate for training the multi-modal fusion network was 1e-3. AdamW \cite{loshchilov2017decoupled} was used as the optimizer. Weight decay was set to 1e-3. Step size of scheduler was 5 during the training of the fusion network and we trained it for 100 epochs. We set $\beta$ to 0.8 in Equation~\ref{eq:joint_loss}. After the fusion network was properly trained, we pre-generated the fusion images, and trained depth estimation network (i.e., Depthformer and SimIPU) using their default settings \cite{lidepthtoolbox2022}. 

\subsection{Evaluation Metrics}
We use standard evaluation protocols following~\cite{eigen2014depth} to evaluate our framework and all baseline methods. Specifically, we measure the mean
absolute relative error (Abs. Rel.), mean squared relative error (Sq. Rel.), root mean squared error (RMSE), and mean log10 error (Log10). Apart from the error metrics, we also adopted three different thresholds as the accuracy metrics which are the common practice in the literature, i.e., $\alpha = 1.25^i, i = 1, 2, 3$.

\subsection{Baseline Methods}
We implement six baselines to compare and examine the effectiveness of our framework on boosting depth estimation, i.e., the use of low-light enhancement and fusion with event and RGB modalities. As mentioned early, salient edges and texture are core features for depth estimation. We therefore adopted the Sobel operator \cite{kanopoulos1988design}, which is an edge detector, to process the RGB modality, and using the resulting image as the alternative to the event image in our framework to justify the use of event data, which is also able to retain salient edge information.

\begin{enumerate}
    \item RGB: the raw RGB image is fed directly into the depth estimation network as the only input for depth estimation.
    \item Event: the event image is fed directly into the depth estimation network as the only input.
    \item RGB + Sobel: the paired raw RGB and Sobel operator processed images are used as the inputs to the phase-2 of EVEN, followed by depth estimation of phase-3. 
    \item RGB + Event: the paired raw RGB and event images are used as the inputs to the phase-2 of EVEN, followed by depth estimation of phase-3.
    \item RGB$_{Enhanced}$: the enhanced RGB image after phase-1 is fed directly into the depth estimation network as the only input for depth estimation.
    \item RGB$_{Enhanced}$ + Sobel: the paired enhanced RGB image after phase-1 and Sobel operator processed image are used as the inputs to the phase-2 of EVEN, followed by depth estimation of phase-3.
\end{enumerate}

\begin{table*}[]
\caption{Results on MonoANC Dataset When the Depth Estimation Network in EVEN is Instantiated as Depthformer and SimIPU Respectively}
\label{tab:overall_results}
\resizebox{\textwidth}{!}{
\begin{tabular}{@{}ccccccccccccccc@{}}
\toprule
\multicolumn{1}{l}{} & \multicolumn{7}{c}{Depthformer}                          & \multicolumn{7}{c}{SimIPU}                                                \\ \cmidrule(l){2-15} 
Input Sequence &
  \multicolumn{4}{c}{Error Metric ↓} &
  \multicolumn{3}{c}{Accuracy Metric ↑} &
  \multicolumn{4}{c}{Error Metric ↓} &
  \multicolumn{3}{c}{Accuracy Metric ↑} \\ \cmidrule(l){2-15} 
                     & Abs. Rel. & Sq. Rel. & RMSE  & Log10 & $\alpha 1$    & $\alpha 2$    & $\alpha 3$    & Abs. Rel. & Sq. Rel.         & RMSE           & Log10 & $\alpha 1$    & $\alpha 2$    & $\alpha 3$    \\ \midrule
RGB                  & 0.192   & 0.310  & 4.973 & 0.069 & 0.810 & 0.911 & 0.985 & 0.293   & 0.370          & 5.177          & 0.079 & 0.710 & 0.921 & 0.972 \\ \midrule
Event                & 0.452   & 0.220  & 7.775 & 0.172 & 0.390 & 0.622 & 0.795 & 0.594   & 1.240          & 9.180          & 0.116 & 0.552 & 0.828 & 0.932 \\ \midrule
RGB + Sobel            & 0.180   & 0.340  &5.304 & 0.064 & 0.808 & 0.908 & 0.956 & 0.266   & 0.310          & 4.947          & 0.067 & 0.773 & 0.930 & 0.976 \\ \midrule
RGB + Event            & 0.179   & 0.340  & 5.992 & 0.067 & 0.795 & 0.920  & 0.956 & 0.229   & 0.280          & 5.151          & 0.057 & 0.837 & 0.953 & 0.984 \\ \midrule
RGB$_{Enhanced}$                 & 0.181   & 0.390  & 5.737 & 0.074 & 0.765 & 0.924 & 0.971 & 0.263   & 0.300          & 4.998          & 0.058 & 0.824 & 0.948 & 0.984 \\ \midrule
RGB$_{Enhanced}$ + Sobel            & 0.139   & \textbf{0.280}  & 5.023 & 0.063 & 0.806 & 0.970 & 0.988 & 0.216   & \textbf{0.240} & \textbf{4.080} & 0.063 & 0.846 & 0.954 & 0.986 \\ \midrule
EVEN (Ours) &
  \textbf{0.112} &
  \textbf{0.280} &
  \textbf{4.335} &
  \textbf{0.049} &
  \textbf{0.903} &
  \textbf{0.976} &
  \textbf{0.993} &
  \textbf{0.125} &
  0.280 &
  4.845 &
  \textbf{0.049} &
  \textbf{0.857} &
  \textbf{0.959} &
  \textbf{0.988} \\ \bottomrule
\end{tabular}
}
\end{table*}

\begin{table*}[]
\caption{Cross Validation Results of EVEN on Different Adverse Weather Conditions}
\label{tab:cross_validation}
\resizebox{\textwidth}{!}{
\begin{tabular}{@{}cccccccccccccccc@{}}
\toprule
\multicolumn{2}{c}{\multirow{2}{*}{Input Sequence}}           & \multicolumn{7}{c}{Depthformer}                                                                                      & \multicolumn{7}{c}{SimIPU}                                                                                           \\ \cmidrule(l){3-16} 
\multicolumn{2}{c}{}                                          & \multicolumn{4}{c}{Error Metric ↓}                                  & \multicolumn{3}{c}{Accuracy Metric ↑}              & \multicolumn{4}{c}{Error Metric ↓}                                  & \multicolumn{3}{c}{Accuracy Metric ↑}              \\ \midrule
Train Set                     & Test Set                      & Abs. Rel.      & Sq. Rel.       & RMSE           & Log10          & $\alpha 1$             & $\alpha 2$             & $\alpha 3$             & Abs. Rel.      & Sq. Rel.       & RMSE           & Log10          & $\alpha 1$             & $\alpha 2$             & $\alpha 3$             \\ \midrule
rain and fog at the same time & rain only and fog only        & 0.325          & 1.987          & 8.475          & 0.187          & 0.471          & 0.645          & 0.797          & 0.330          & 1.865          & 8.710          & 0.187          & 0.420          & 0.655          & 0.786          \\ \midrule
rain only and fog only        & rain and fog at the same time & \textbf{0.267} & \textbf{0.315} & \textbf{4.934} & \textbf{0.031} & \textbf{0.646} & \textbf{0.833} & \textbf{0.937} & \textbf{0.260} & \textbf{0.307} & \textbf{4.933} & \textbf{0.031} & \textbf{0.680} & \textbf{0.844} & \textbf{0.939} \\ \bottomrule
\end{tabular}}
\end{table*}

\subsection{Overall Results}

As we instantiate the depth estimation network separately as either a Depthformer or a SimIPU, we run six baselines accordingly based on the instantiated depth estimation network. Table~\ref{tab:overall_results} summarizes the overall results. Our complete EVEN framework outperforms the baseline methods, and its performance improvement is consistent across Depthformer and SimIPU. The absolute relative error (Abs. Rel.) is reduced by 41.7\% and 57.3\% respectively compared to a single RGB input to the Depthformer and SimIPU. An 11.5\% relative improvement on $\alpha 1$ accuracy metric can also be observed for EVEN using Depthformer, and a 20.7\% increase for EVEN using SimIPU, compared to a single RGB image as the input to these two depth estimation networks.

\begin{figure*}[]
\centering
\centerline{\includegraphics[width=\textwidth,scale=1.0]{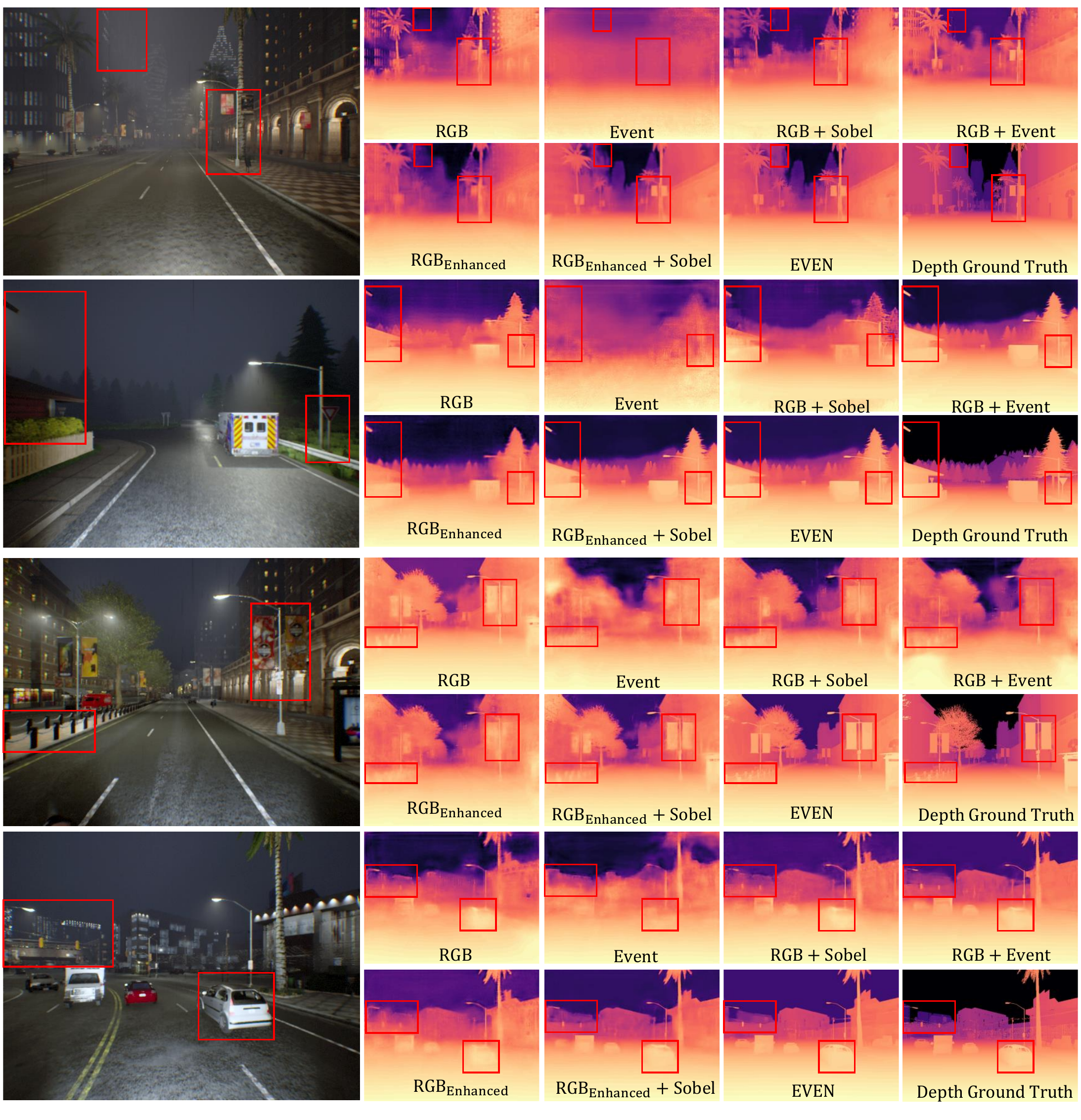}}
\caption{Qualitative results of depth estimation on MonoANC dataset. Top two examples are the results when Depthformer is adopted as the depth estimation network, and the bottom two examples are the results when SimIPU is adopted as the depth estimation network. Areas indicated by the red boxes show that our EVEN framework can better estimate monocular depth than other baseline methods.}
\label{fig:qualitative_results}
\end{figure*}

Fig.~\ref{fig:qualitative_results} shows four qualitative results of depth estimation on MonoANC. It can be observed that the depth maps estimated by EVEN have a noticeable improvement in detail at the edges as well as the objects in the far distance compared to those of baselines. As indicated by the red boxes in Fig.~\ref{fig:qualitative_results}, our complete EVEN framework can produce depth maps without much artifacts, and are closer to the ground truth. These visually prove that the fusion of the edge information and HDR features of event data in EVEN is effective. When we replace the event image with Sobel operator processed image, i.e., indicated by RGB$_{Enhanced}$ + Sobel, the quality of the estimated depth map slightly degrades, but is still better than those of the rest baseline methods.

\subsection{Cross Validation on Adverse Weather}
We further split MonoANC based on different weather conditions. Specifically, there are three adverse weather conditions as shown in Fig.~\ref{fig:dataset_statistics}(b): 1) rain only; 2) fog only; 3) rain and fog occur together. We split the dataset into two sets. One set contains samples of rain only and fog only, and the other set contains samples of simultaneous occurrence of rain and fog in the scene. A two-fold cross-validation is then conducted to evaluate the performance of EVEN. Table~\ref{tab:cross_validation} shows the results. When the framework has seen each individual weather condition during the training, it can well estimate the depth of the scene with mixed adverse weather conditions, i.e., rain and fog occurring at the same time in the scene. Conversely, it becomes difficult for the framework to estimate depth for the scenes with only a single adverse weather condition if the training data is scenes of mixed adverse weather. Hence, cost function of decomposing adverse weather combinations is worth investigating for better depth estimation in future work.

\section{Conclusion}\label{sec:conclusion}
In this paper, we have proposed a framework that integrates low-light enhancement and fuses RGB and event modailies for effective monocular depth estimation under adverse night conditions. A synthetic night-time driving dataset that contains paired RGB, event and depth images has also been constructed, which includes scenes that encountering adverse weather, light and road conditions. The experiment results have shown that our proposed framework is able to achieve satisfactory depth estimation results in various adverse night scenarios.



\end{document}